# Effects of Leader's Position and Shape on Aerodynamic Performances of V Flight Formation

**H.P.Thien, M.A.Moelyadi,** and **H. Muhammad**

Aeronautics and Astronautics Department
Bandung Institute of Technology
Ganesha 10, 40132, Bandung, Indonesia
e-mail: moelyadi@ae.itb.ac.id
Phone:+62 – 22 - 2504529

**Abstract**

The influences of the leader in a group of V – flight formation are dealt with. The investigation is focused on the effect of its position and shape on aerodynamics performances of a given V – flight formation. Vortices generated the wing tip of the leader moves downstream forming a pair of opposite rotating line vortices. These vortices are generally undesirable because they create a downwash that increases the induced drag on leader's wing. However, this downwash is also accompanied by an upwash that can beneficial to the follower's wing flying behind the leader's one, namely a favorable lift for the follower's wing. How much contributions of the leader's wing to the follower's wing in the V – formation flight is determined by the strength of tip vortices generated by the leader's wing which is influenced by its position and shape including incidence angle, dihedral angle, aspect ratio and taper ratio.

The prediction of aerodynamic performances of the V – flight formation including lift, drag and moment coefficients is numerically performed by solving Navier – Stokes equations with k - ε turbulence model. The computational domain is defined with multiblock topology to capture the complex geometry arrangement of the V – flight formation.

**Nomenclature**

| | |
|---|---|
| $L$ , $L'$ | lift and the rotation of original lift |
| $D$ , $D'$ | drag and the rotational of original drag |
| $P$ | power |
| $C_L$ , $C_D$ | coefficient of lift and drag |
| $\alpha$ | angle of attack |
| $b$ | wing span |
| $\Delta$ | change in parameter |
| $WTS$ | wing tip spacing |
| Subscript | |
| $FF$ | formation flight |
| $BL$ | baseline |

## 1 Introduction

Formation flight can result in large induced drag reductions. By doing the V – shape formation consisting of 25 members, each bird can achieve a reduction in induced drag as large as 65% that could result in a range increase about 70% [1]. The full advantage of bird formation flying can be obtained by combining the variability of bird's wings and their highly controllable flight, which allow them to change their wing geometry as well as fly very close together.

The leader of an acute V formation saves less energy than the trailing participants do. The disadvantage of leading is reduced in more obtuse formations, and when the longitudinal distance between neighbor is small [2].

This paper presents an analysis of the effect of the leader to the aerodynamics performance of V – flight formation consisting of three members using the numerical simulation method, computed on simple wing model. The first section of the paper will focus on review of previous works on formation flight research as well as on the basic theory of this favor aerodynamic interference phenomena.

The computational aerodynamics of wings in formation flight will be dealt with in the second section of this paper including discretized the computational domain and solve the governing flow equation. Then, result analyses together aerodynamics characteristics are described.

## 2 Background

### 2.1 State of art

Formation flight has been studied frequency in the past several decades. There have a numbers of researchers interested in the bird as well as airplane formation flight, and they have had various degree of success.

In the early of 1970s, Schollenberger and Lissaman investigated the formation flight of bird; their analysis showed that induced drag reduction of each bird in an arrow formation could reach the number of 45%, compared with another one in solo flight [1]. Hummel [3] derived a theory of optimal wing tip spacing to obtain the maximum reduction in induced power required, the optimum wing tip spacing that maximizes induced power saving will be a negative value, which can only be achieve in the V formation. Beukenberg and Hummel [4] gave flight test data for three aircraft flying in an arrow formation, and compared it to the theoretical aerodynamic results obtained with various vortex models.

Recently, formation flying has again become of interest. At Virginia Polytechnic Institute and State University, Iglesias, S. carried out the study on investigating the effect of relative distance in the three dimensions to the induced drag reduction and found out the corresponding spanloads for a group of aircraft in an arrow formation [5].

Formation flight theory has been validated by analytical studies and recent autonomous formation flight. For example, Fowler, J.M. and Andrea, R.D. [6] conducted an experiment on large formation of 31 wings, the result showed that induced drag reduction was as much as 41%. NASA Dryden Research Center did the formation flight test on two F/A – 18 aircrafts which have the same configuration. Significant performance benefit have been obtained from the flight test phase, such as drag reduction of more than 20% and fuel flow reduction of more than 18% has been measured at flight condition of Mach 0.56 and altitude of 25000ft [7].

### 2.2 Basic theory

The dominant aerodynamic forces that affect flight are lift and drag. The difference in the pressure of air above and below a wing produces lift. When a wing is hold at a sight angle to an air current, air flows faster over the upper surface than it does over the lower surface, thus creating less pressure above the wing than below it and causing lift. At the same time, drag, or resistance to the moving air, drags the wing backward. The combined effect of these two forces lifts the wing and drags it backward.

The pressure difference between the upper and lower surface also causes a spanwise flow of air outward toward the tips on the lower surface, around the tips and inward toward the center of the wing, this motion creates what is called a wingtip vortex. Combined with the free stream velocity, this spanwise flow produces a swirling motion of the air trailing downstream of the wing. This motion is referred to as the wing's trailing vortex system.

Immediately behind the wing, the vortex system is shed in the form of a vortex sheet, which rolls up rapidly within a few chord lengths to form a pair of oppositely rotating line vortices. Looking in the direction of flight, the vortex from the left wing tip rotates in a clockwise direction; the right tip vortex rotates in the opposite direction.

Figure 1 shows the vortices system in a finite wing. The principle effect of the trailing vortex system is that it reduces the angle of attack of each section by a small decrement known as the induced angle of attack $\alpha_i$. The lift vector for a wing section is seen to be titled rearward through the induced angle of attack. This component, integrated over the wingspan, results in the induced drag.

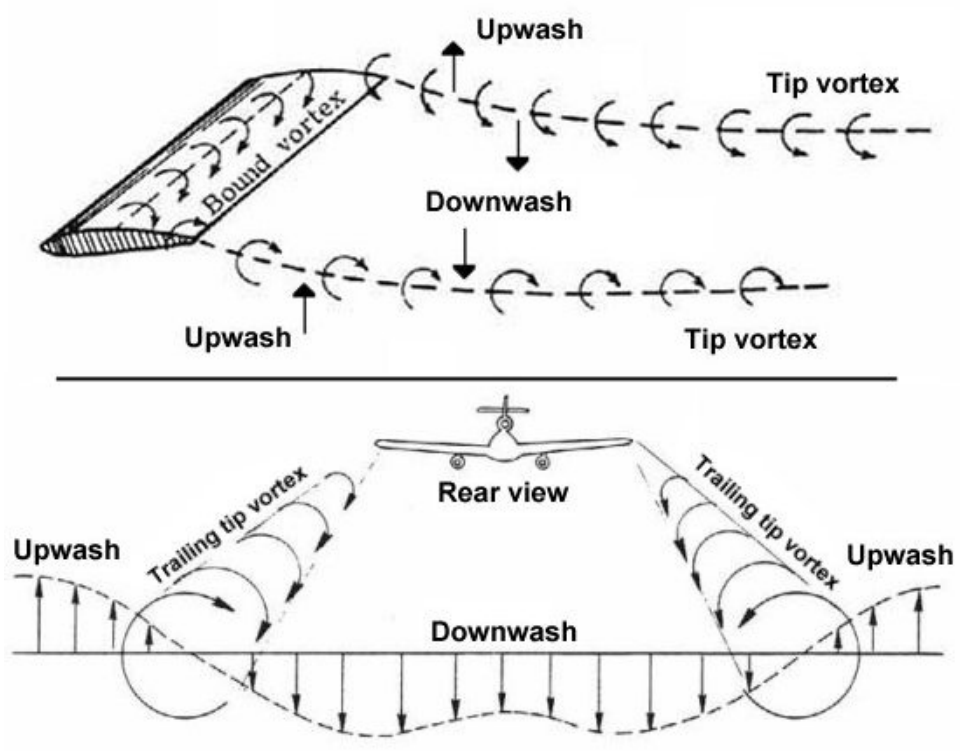

**Figure 1: Regions of upwash and downwash created by trailing vortices**

Vortices are generally undesirable because they create a downwash that increases the induced drag on a wing in flight. However, this downwash is also accompanied by an upwash that can be beneficial to a second wing flying behind and slightly above the first.

The common theory on formation flight states that drag reduction is actually attained because of the rotation of the lift vector that occurs while the rear wing is in the effect of the front airplane [7].

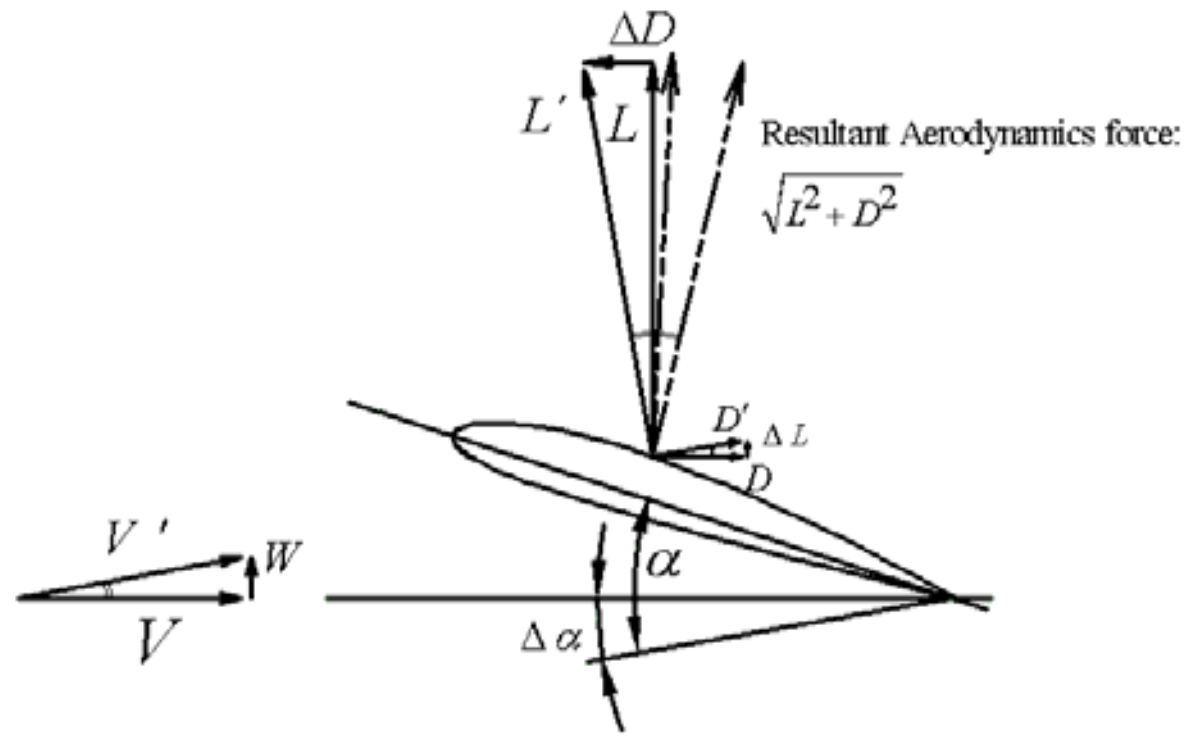

**Figure 2: Rotation of resultant force cause by upwash of the leading airplane**

Figure 2 show the rotation of resultant aerodynamic force caused by upwash of the leading wing. The rotation of lift and drag value are presented by $L'$ and $D'$. Primary assumption in this derivation is that lift is much greater than drag and the increase of angle of attack $\Delta\alpha$ is sufficiently small enough so that the approximations in trigonometry can be used with a great level of accuracy [7].

Due to the traditional bookkeeping methodology, the actual lift and drag values are maintained relative to the vehicle's global, rather than local flight path during formation flight. As a result, the resultant aerodynamic force remains constant because the upwash only rotates this force.

$$\sqrt{L^2 + D^2} = \sqrt{L'^2 + D'^2} \qquad (1)$$

The change in lift and drag due to the rotation of the lift force is determined as

$$\Delta L = D\sin\left(\Delta\alpha\right) \quad (2)$$

$$\Delta D = L\sin\left(\Delta\alpha\right) \quad (3)$$

The lift and drag in formation flight is obtained by

$$D_{FF} = D'\cos\left(\Delta\alpha\right) + \Delta D \quad (4)$$

$$L_{FF} = L'\cos\left(\Delta\alpha\right) + \Delta L \quad (5)$$

Because lift tends to be an order of magnitude greater than drag $L \geq D$, so drag is influenced more by the rotation effect than lift

$$L\sin\left(\Delta\alpha\right) \geq D\sin\left(\Delta\alpha\right) \quad (6)$$

Substituting eqs. (2) and (3) into (6), it becomes

$$\Delta D \geq \Delta L \quad (7)$$

A considerable reduction in drag can be obtained by a small upwash angle while an insignificant increase in lift simultaneously occurs.

Percent drag reduction is determined as

$$Percent\ drag\ reduction = \Delta D / D_{BL} = 1 - \left(D_{FF} / D_{BL}\right) \quad (8)$$

Equation for calculating the power reduction is expressed as

$$Percent\ power\ reduction = \Delta P / P_{BL} = 1 - \left(D_{FF} / D_{BL}\right) \quad (9)$$

The flowing paragraph focuses on analysis the effect of the distances between members in an arrow formation flight consists of three aircrafts, including streamwise, spanwise, and vertical distance, on the induced drag reduction.

According to Munk's theory, the change in induced drag coefficient for each aircraft will be changed as the streanwise relative distance is changed but the total induced drag for the formation is not changed, namely the formation induced drag is independent of the streamswise location of the airplane.

Considered for each aircraft, the induced drag coefficient are highly dependent on streamwise position as the aircrafts are close together. For a streamwise distance between two aircrafts greater than three spans, the induced drag in each aircraft reaches a steady value and no longer depends on streamwise separation.

On the aspect of spanwise separation, the formation induced drag is highly dependents on this parameter. The maximum induced drag reduction occurs at a negative wing tip spacing value [8]. However, beyond a critical negative spacing, the following aircraft starts to fly in the central downwash and experiences a negative saving. Conversely, for a wing tip spacing of one span, the drag saving are very small and formation flight is no longer beneficial.

The formation flight achieves the maximum drag reduction value when the aircrafts are on the same horizontal plane, namely no separation in vertical direction.

In summary, the formation induced drag reduction is independent of the streamwise position of aircraft. The maximum reduction is obtained at negative wing tip spacing and zero vertical separation.

According to Hummel [9] the optimum of wing tip spacing that maximizes the induce power saving will be a negative value, expressed as

$$WTS_{optimum} = 0.5b\left(1 - 0.89\right) \quad (10)$$

## 3 Numerical Simulation Method for Formation Flight Aerodynamics Analysis

This section describes two main steps of the numerical simulation method that is used to find the solution for formation flight problems in this research; domain discretization and flow field properties evaluation. The dicscretization of domain of interest is performed by generating mesh using the available grid generator. Then, flow field properties are obtained by solving the Navier – Stokes equations with k - ε turbulence model.

### 3.1 Formation flight model definition

The following paragraph depicts clearly the formation flight models that used in this research.

Each bird in a V – shape formation flight consisting of three members will be treated as a simple wing. Different leading wing configuration is used while the two same trailing wings are kept intact. All of wings use the same airfoil.

Table 1. The characteristics of the trailing wings

| | |
|---|---|
| Airfoil | NACA9405 |
| Span | 1.6m |
| Chord | 0.16m |
| Aspect ratio | 10 |
| Taper ratio | 1 |
| Sweep angle | 0deg |

The same configuration will be used for the whole wings in the investigation of the effect of incidence and dihedral angle of the leader to the aerodynamics performance. Three cases of incidence angle; 0, 5, 10 will be dealt with for analyzing the effect of incidence angle. The effect of leader's position in terms of dihedral angle is conducted by doing the computational simulation on formation model with different leading wing dihedral angle, which is 0deg, 20deg and 45deg.

An aspect ratio of 10 together with an aspect ratio of 6, which is obtained by varying only the span are used for leading wing model in the study of leader' shape effect in terms of aspect ratio.

In the other case of leader's shape effect, three values of taper ratio; 0, 0.3 and 0.5 are obtained by varying the chord along the span, that is the same with the trailing wing's.

One the wings configuration is established, it is necessary to consider the relative position between wings in formation group. This relative position is kept to be same for all of formation cases mentioned above. The arrangement of formation is symmetric, illustrated in figure 3. Distance between wings in three directions are showed in table 2, these distance includes:

- Streamwise distance: distance between the trailing edge of the front wing to leading edge of the rear wing, measured in the streamwise direction.
- The spanwise distance: the spacing between two wing tips, measured in the spanwise direction.
- Vertical distance, preferred to the distance between sections at the middle of wings, measured in the vertical direction.

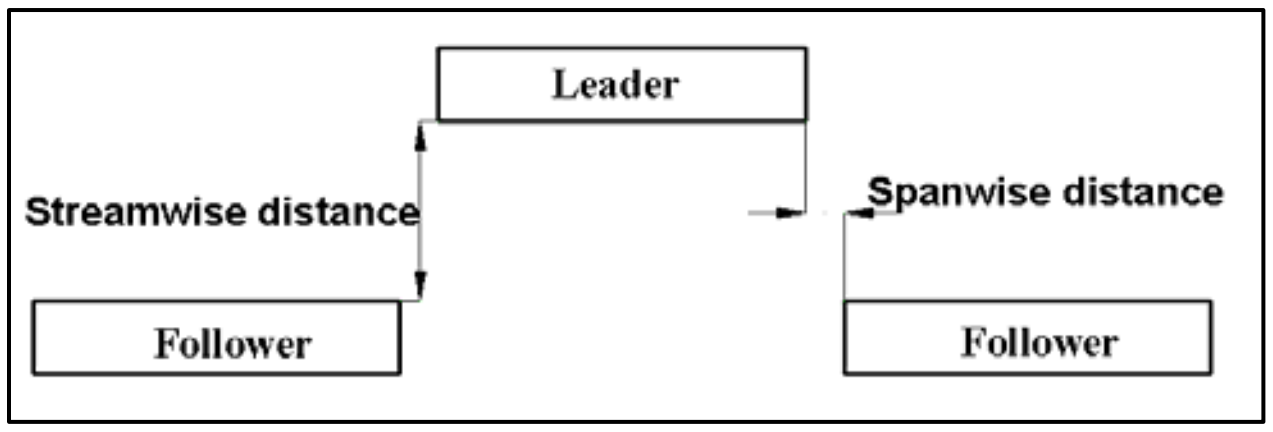

**Figure 3: Geometry arrangement in V formation**

Table 2. Wings relative position

| Stream wise | Spanwise (wing tip spacing) | Vertical |
|---|---|---|
| 0.16 m (1C) | 0 | 0 |

**3.2 Topology definition**

The procedure of mesh definition in computational domain is initiated with the geometry definition. Then, a initial block is obtained based on a suitable topology. Multi-block technique together with modificative tool such as splitting block, merging the block , O – grid block are used to achieve a more precise block topology with the complex geometry arrangement of V formation flight. The topology and blocks of the basic case of V formation is presented in figure 4.

Distribution of points along the edges of the block is carried out as the prepared step for mesh generation. A significant numbers of points are distributed in critical regions such as on leading edge, trailing edge, streamwise and spanwise section of the geometry. The point distribution is also different for these above mentioned sections, it becomes denser when the points approaching the body surface. The mesh generation for the wings in V – formation is exhibited in figure 5.

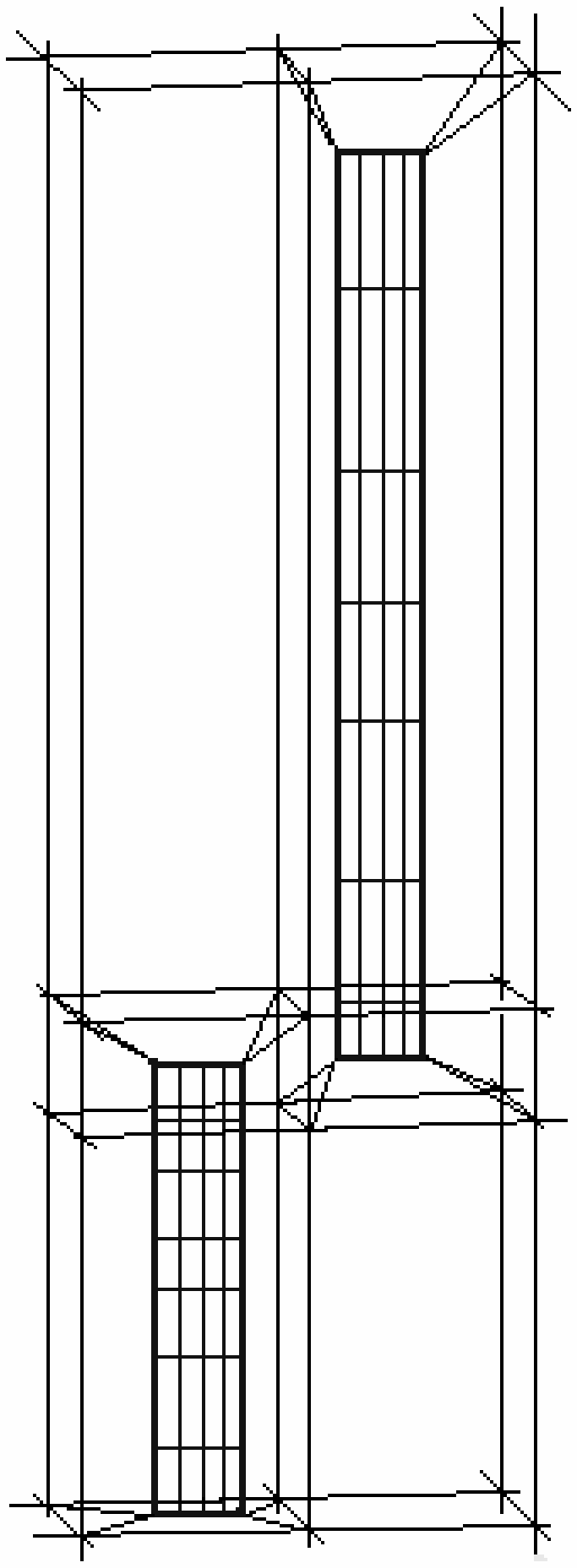

**Figure 4: Topology of formation flight model**

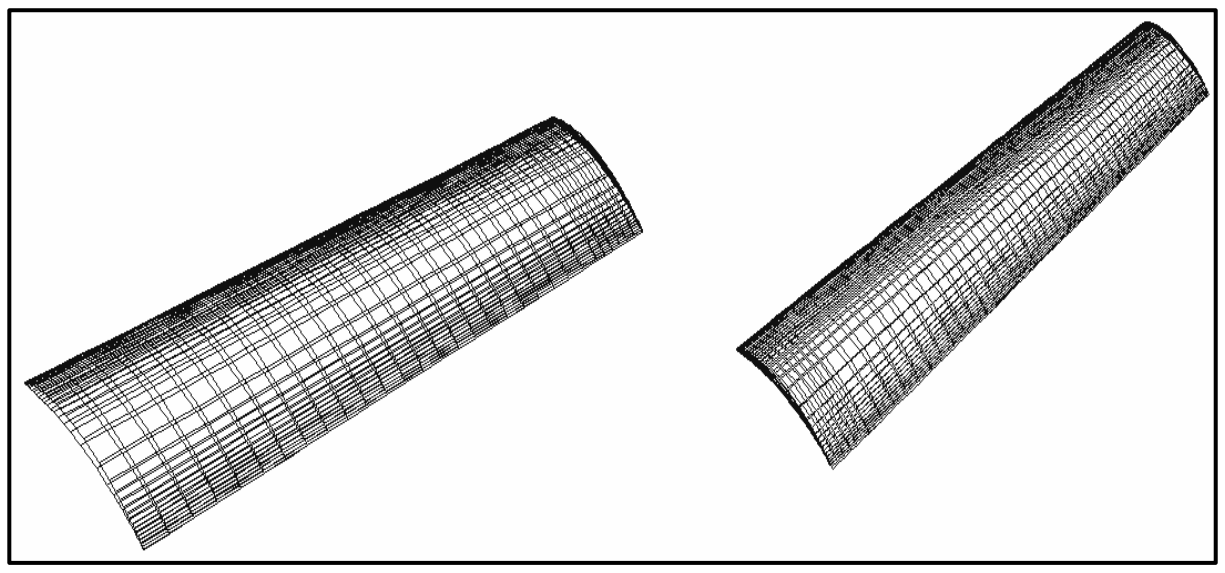

**Figure 5: Mesh generation for wings in V formation model**

**3.3 Solver definition**

The solver based on solution of the Navier – Stokes equations with k - ε turbulence model is used to obtain the flow field properties.

## 4 Results and Discussions

Lift, drag and moment of each member in a group of three wings flying in an V – shape formation will be found, and these coefficients will be compared as a function of shape and position of the leading wing, including incidence angle, aspect ratio, dihedral and taper ratio.

### 4.1 Incidence angle and Aspect ratio effect

The incidence angle $i$, showed in figure 6 will be varied from 0 to 5 and 10deg, the variation in lift and drag saving will be studied as a functions of this angle.

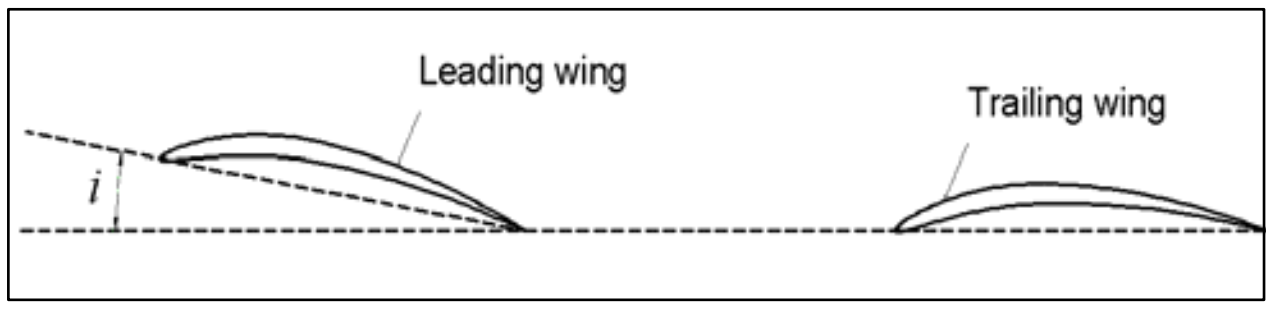

**Figure 6: Incidence angle between two wings**

Figure 7, 8 and 9 shows the change in total lift, drag and pitching moment for each wing in formation as the relative incidence angle is changed. The results are shown as the ratio of these values in the formation to their corresponding values when flying a lone.

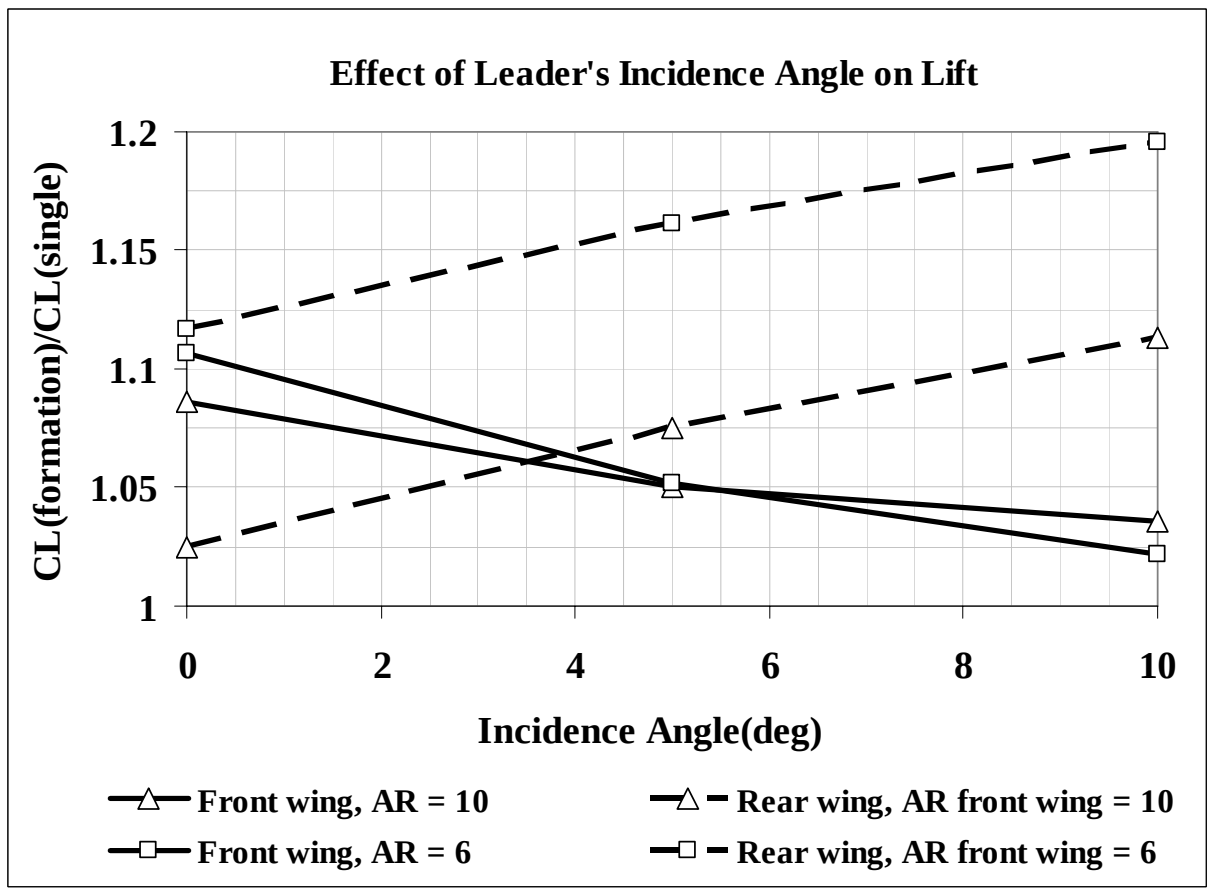

**Figure 7: Effect of leader's incidence angle on lift**

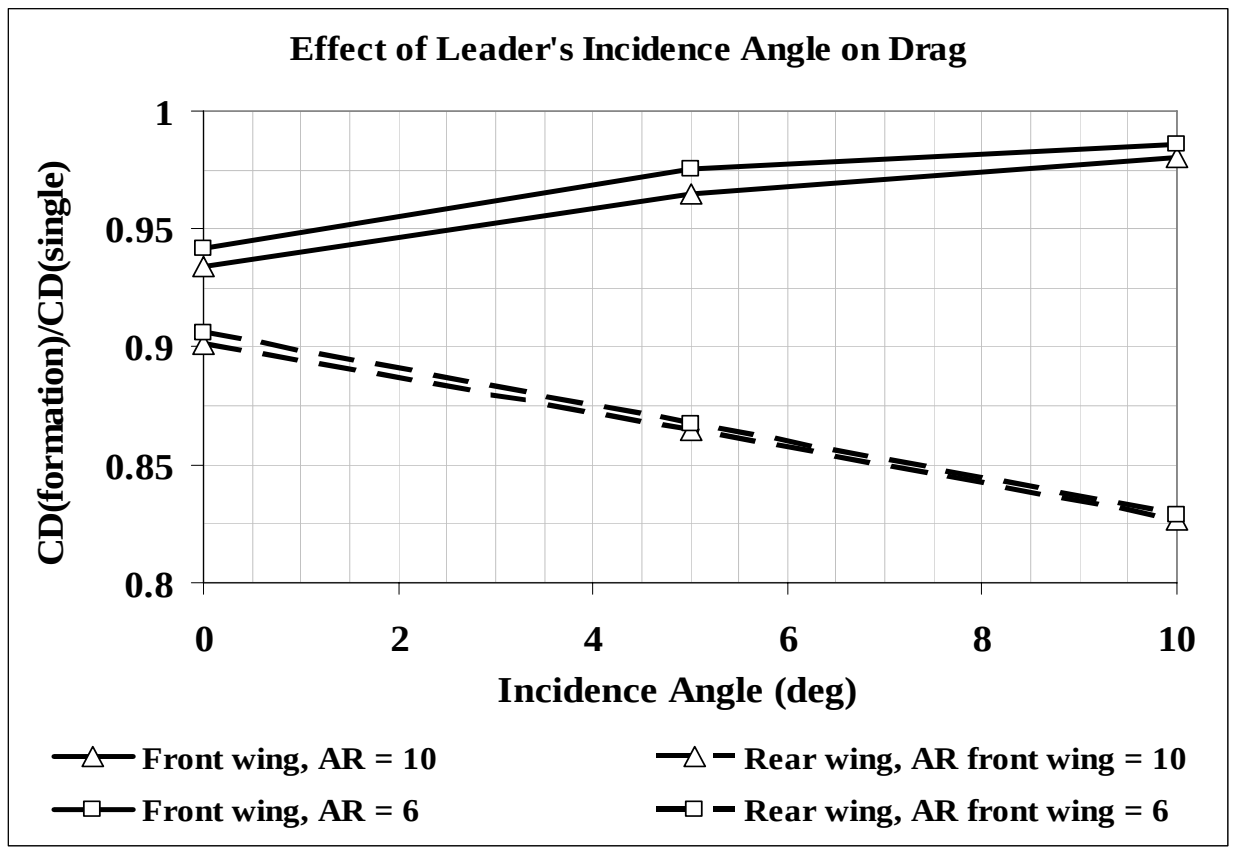

**Figure 8: Effect of leader's incidence angle on drag**

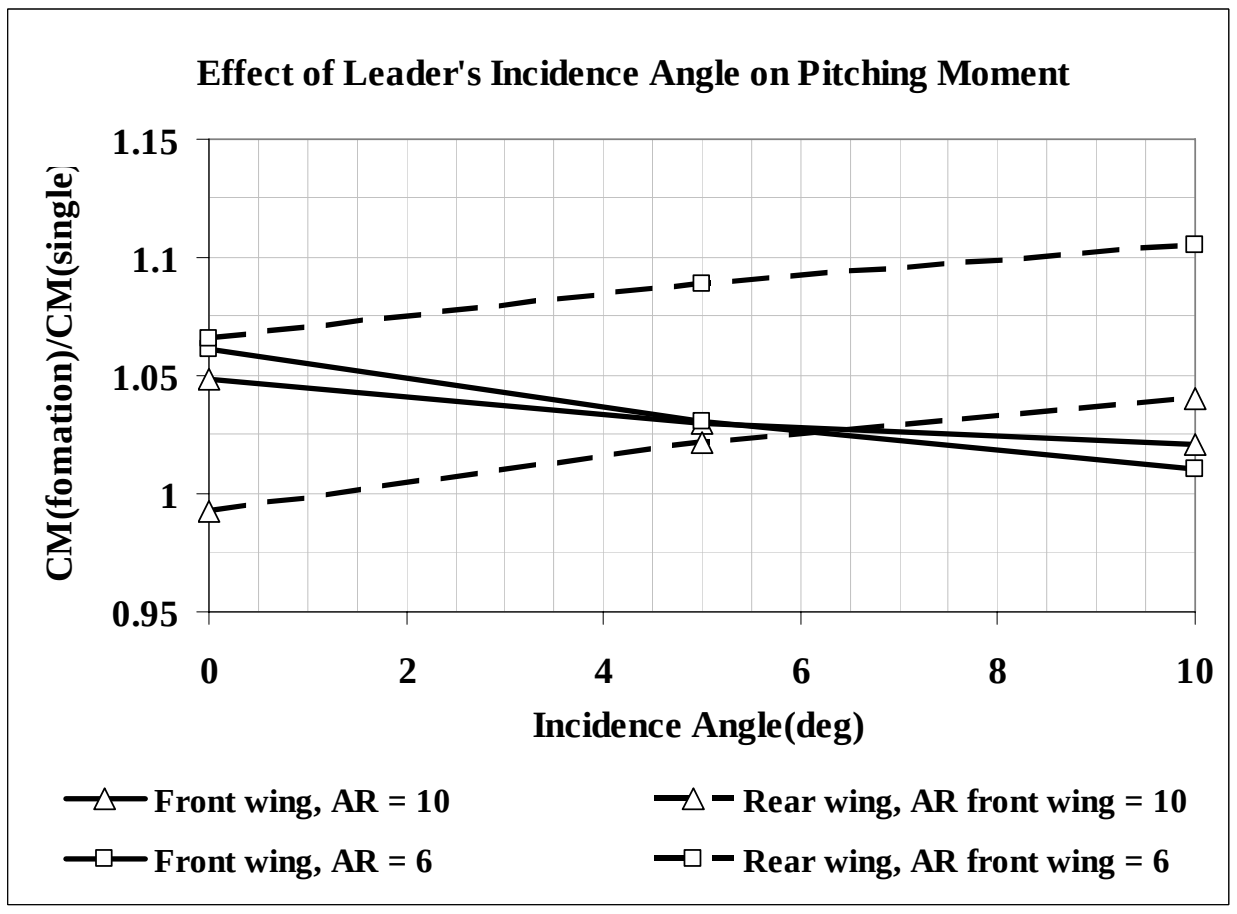

**Figure 9: Effect of leader's incidence angle on pitching moment**

Wings in formation get the benefit of increase in lift due to the effect of the leader's upwash. As a result, a wing flied in a formation group will able to get a higher lift compared with a same wing in alone flight. The lift increase in trailing wing is directly proportional to the incidence angle of the leader, while the increase lift in the leading wing tends to decrease with the increase of its incidence angle.

Aspect ratio also influences the lift increase; a low aspect ratio leading wing leads to a high lift increase in trailing wings. The difference in lift increase can achieve up to 10% for a wing has aspect ratio of 10 compared with a wing with aspect ratio of 6 at the same angle of incidence.

In terms of drag saving, the aspect ratio nearly has a weak influence. An inappreciable different between drag ratio on the rear wing with different front wing aspect ratio at the same incidence angle can be observed in the figure 8. Maximum value of this difference is about 0.5%. On the contrary, the incidence angle of the leading edge is highly effects on the saving of drag in the rear wing. The effect is taken place in the trend that reduces the drag ratio of the wing in formation and wing in alone flight as the incidence angle of the leader increases, it means that rear wing will get a higher drag saving as increase the leader's incidence angle.

The pitching moment on rear wing of formation, also caused by the lift and drag, tends to increase as the leading wing's incidence angle increase because the prevailingness of lift increase over the drag reduction tends to increase.

The effect of aspect ratio on change of rear wing pitching moment is significant, as the leader's aspect ratio decreases the three-dimension effect on lift and drag increase.

In summary, aspect ratio of the leader affect the change of lift and moment on rear wing of formation, its effect to the drag reduction is insignificant. Leading wing's incidence angle has a highly influence on the change of lift, drag and pitching moment.

### 4.2 Dihedral angle effect

The dihedral angle of the leading wing is varied from 0 to 20 and 45deg while the rear wings' dihedral angle are kept to zero. Figure 10, 11 and 12 shows the effect of dihedral angle of the leading wing on the wings in formation in terms of lift, drag and pitching moment.

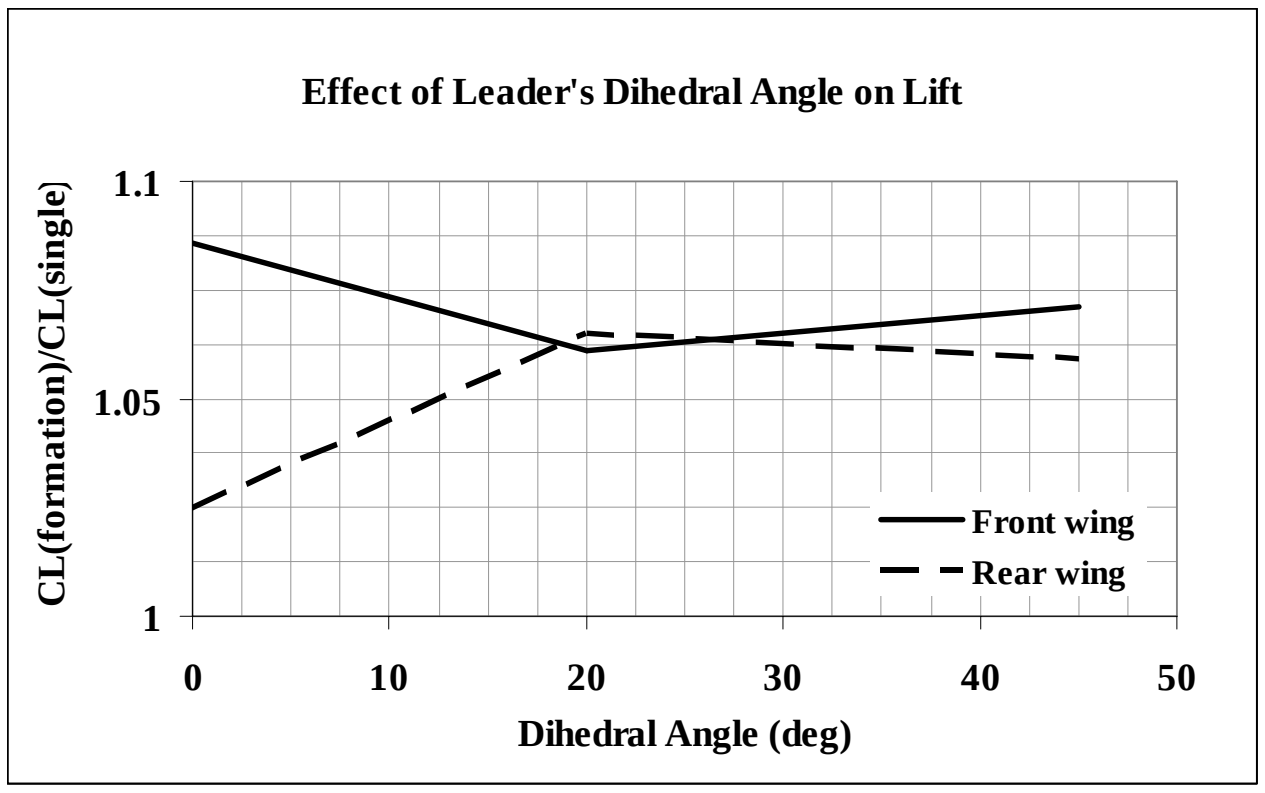

**Figure 10: Effect of leader's dihedral angle on lift**

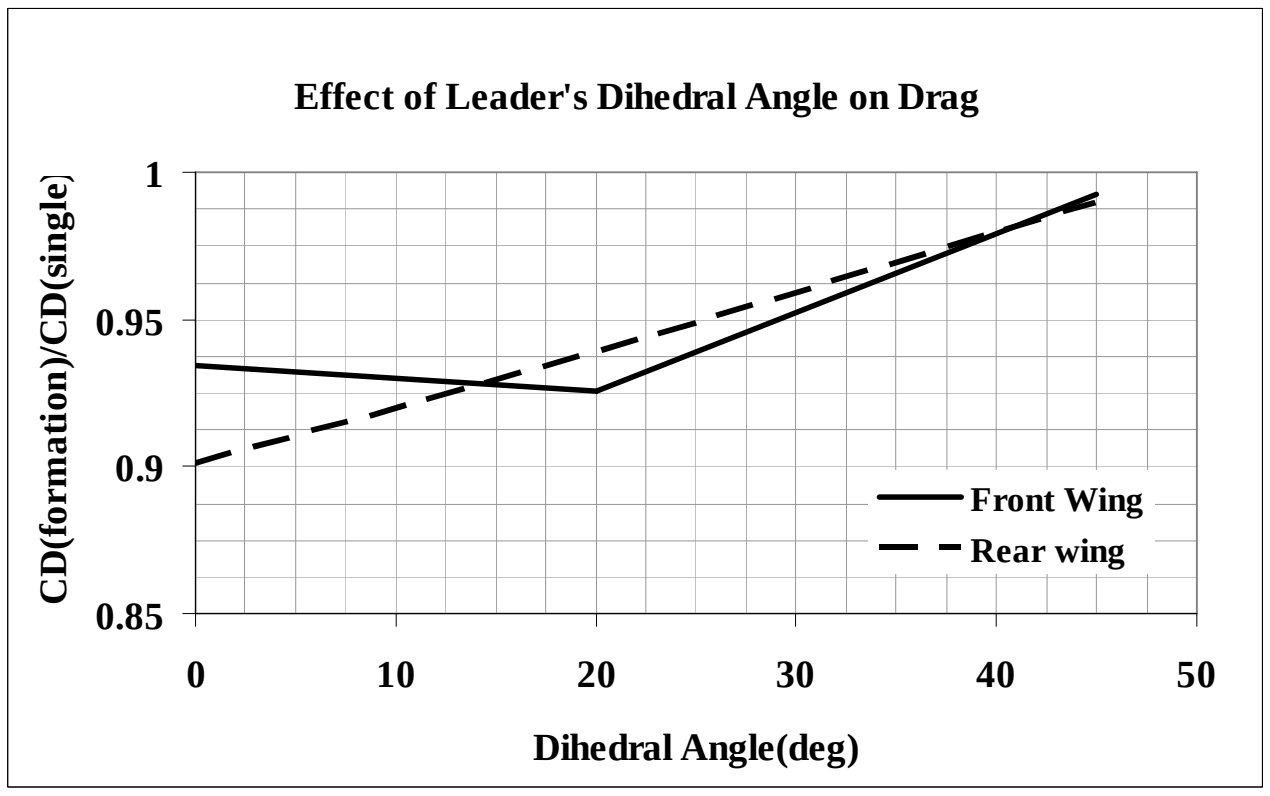

**Figure 11: Effect of leader's dihedral angle on drag**

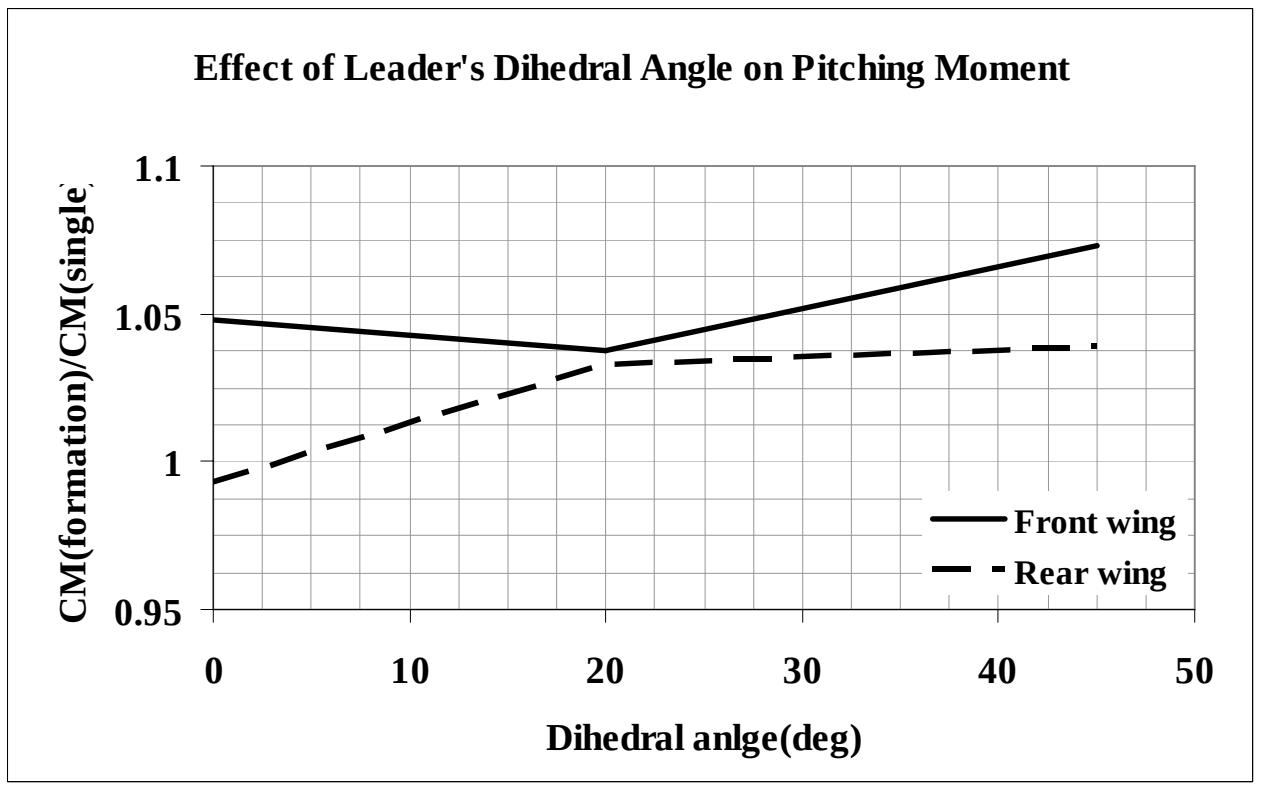

**Figure 12: Effect of leader's dihedral angle on pitching moment**

Figure 10 shows the behavior of the lift ratio of rear wing in formation and in alone flight to the dihedral angle of the leader. This ratio increases considerably at low value of dihedral angle, and changes a little, as the angle is higher.

In the respect of drag reduction, the leader's dihedral angle acts an adverse effect on the amount of drag saving in the rear wing of formation. Rear wing drag reduction seems to linear decrease as the dihedral angle increase. At high value of dihedral angle, the effect of the leader on the rear wings in term of drag reduction no longer exists; the drag on both leading wing and rear wing is nearly same with their value as flying alone.

Figure 12, presents the change in pitching moment ratio as the dihedral angle of the leader is varied. On rear wing, pitching moment ratio increases significant at low leader's dihedral angle. As the angle is higher, the rear wing nearly maintains its pitching moment ratio.

### 4.3 Taper ratio effect

Leading wing with different taper ratio, namely, 0, 0.3 and 0.5 are used to survey the effect of leader shape on the aerodynamic forces and moment of wings in V – formation.

The results of this survey are shown in figure 13, 14 and 15 as the ratio between the lift, drag and pitching moment on leading wing and trailing wing in V – formation to the corresponding ones in alone flight.

Figure 13 shows that at low taper ratio the ratio in lift of rear wing in formation and in alone flight increase as the taper ratio increase. For high taper ratio, the rear lift ratio tends to decrease. Lift ratio is smallest as the leader has the rectangular platform (taper ratio of 1).

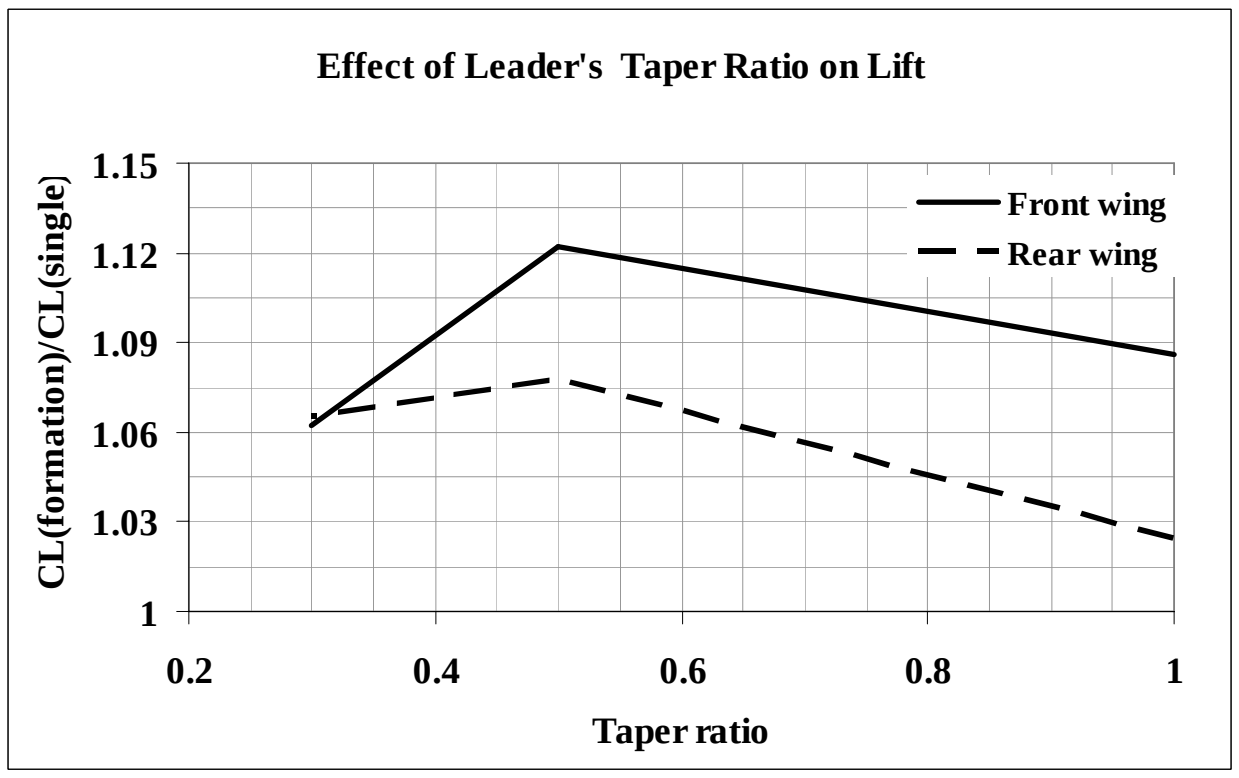

**Figure 13: Effect of leader's taper ratio on lift**

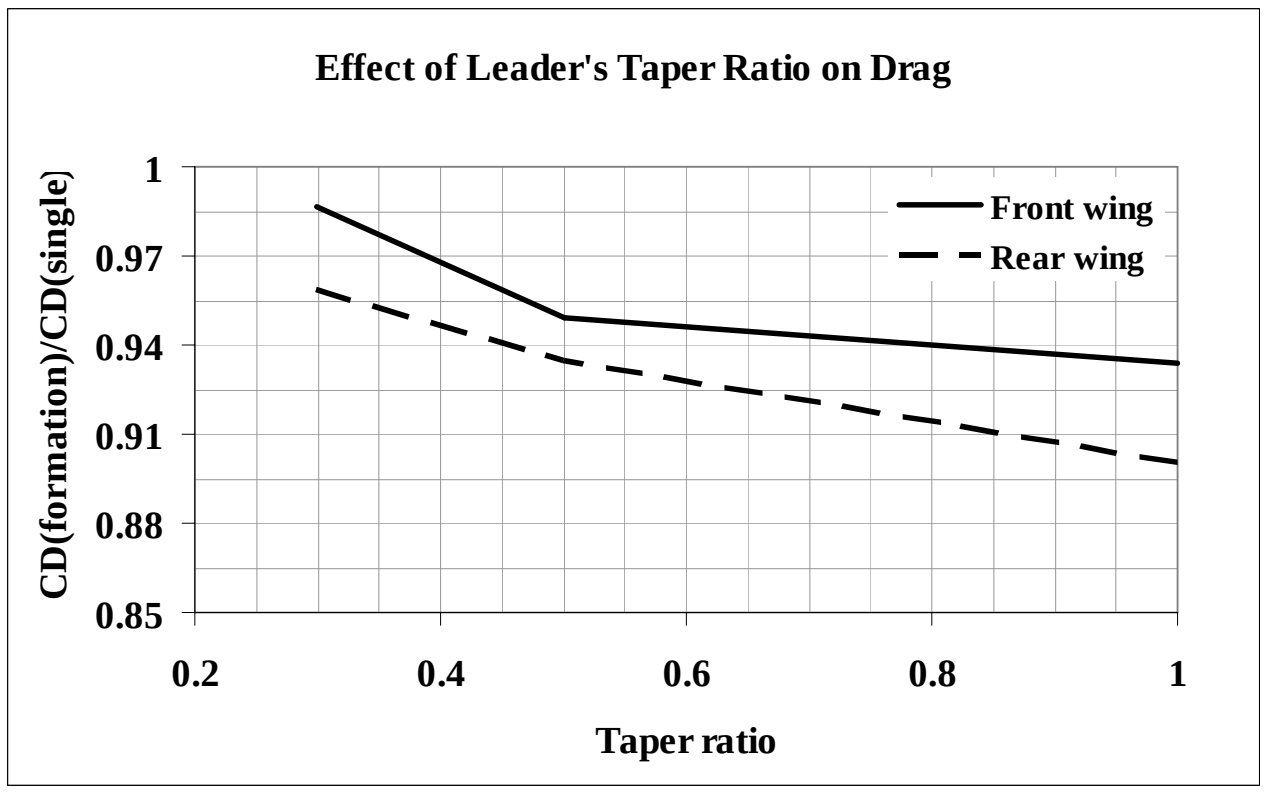

**Figure 14: Effect of leader's taper ratio on drag**

The increase of leader taper ratio has positive effect to the drag saving of the rear wing in formation. As a result, rectangular platform of the leading wing will give the highest drag reduction in rear wing.

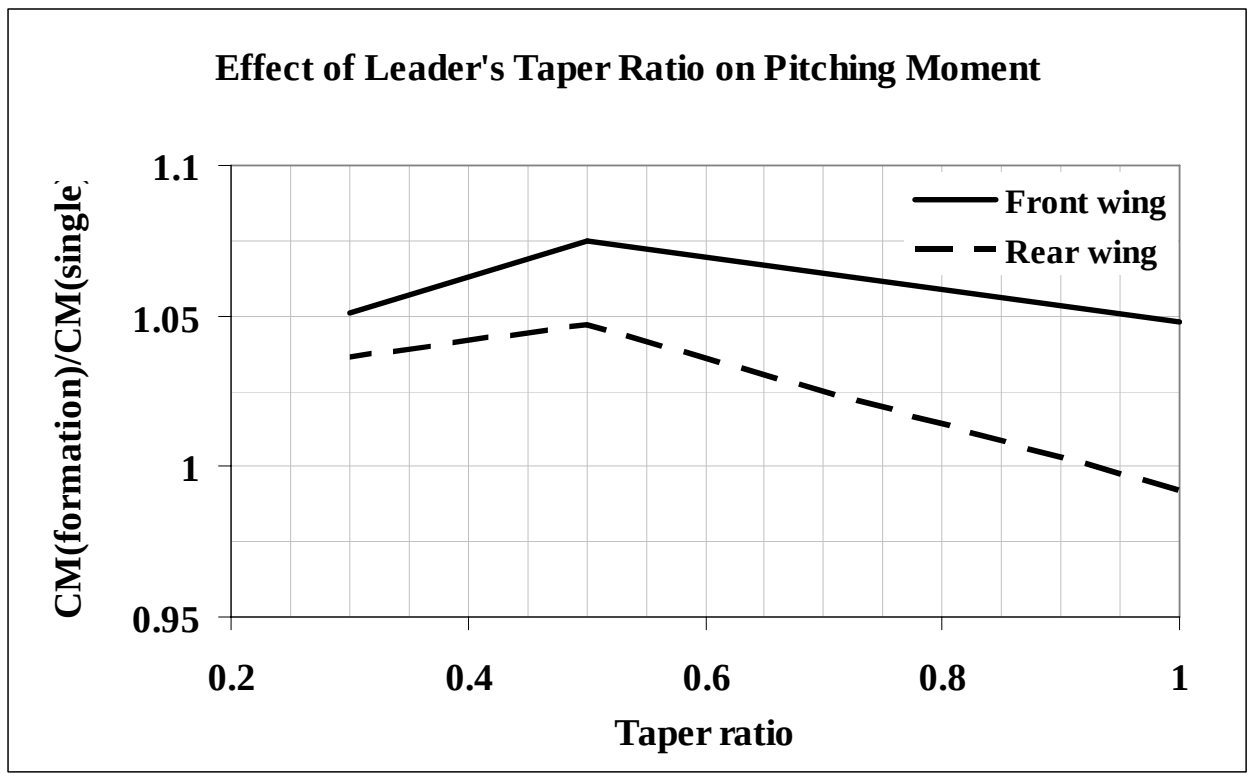

**Figure 15: Effect of leader's taper ratio on pitching moment**

The behavior of the pitching moment on the rear wing in V formation is the same with the behavior of lift as the leader's taper ratio is varied. Pitching moment ratio on rear wing reduces significant, as taper ratio is higher.

## 5 Conclusions

Investigation on the effects of the leader's shape and position on aerodynamics performance of V formation is conducted by using the numerical simulation method.

Leader wing with variation in shape and position parameter including incidence angle, dihedral angle, aspect ratio and taper ratio is used to find out the behavior of lift, drag and moments on the same rear wing.

Some conclusions can be listed as follows:

- Incidence angle has a positive effect on increasing the lift, pitching moment, and reducing the drag and of the rear wing.
- Aspect ratio has a significant effect on increasing the rear wing lift and pitching moment, but seems to be no effect to the drag reduction.
- Dihedral angle has an adverse effect on drag saving in the rear wing. The change in lift of the rear wing is nearly constant at high value of dihedral angle.
- Taper ratio is positive effect to the drag reduction, and its effect on the change of lift is also considerable.
- The behavior of lift and pitching moment are nearly the same in all cases of this investigation.